\documentclass[11pt]{article}

\usepackage[preprint]{acl}

\usepackage{times}
\usepackage{latexsym}
\usepackage{amsmath}
\usepackage{hyperref}
\usepackage[T1]{fontenc}

\usepackage[utf8]{inputenc}

\usepackage{microtype}

\usepackage{inconsolata}

\usepackage{graphicx}
\usepackage{caption}
\usepackage{subcaption}
\title{Language Models Struggle to Use Representations Learned In-Context}

\author{
Michael A. Lepori$^{\dagger}$\thanks{Corresponding author: \texttt{michael\_lepori@brown.edu}; Work done while at Google DeepMind.} \And
Tal Linzen$^{\P\S}$ \And
Ann Yuan$^{\ddagger}$ \And
Katja Filippova$^{\ddagger}$
\AND
{\normalfont
$^{\dagger}$ Brown University 
} \And
{\normalfont
$^{\P}$ Google Research
} \And
{\normalfont
$^{\S}$ New York University
} \And
{\normalfont
$^{\ddagger}$ Google DeepMind
}
}

\begin{document}
\maketitle
\begin{abstract}
Though language models (LMs) have enabled great success across a wide variety of tasks, they still appear to fall short of one of the loftier goals of artificial intelligence research: creating an artificial system that can adapt its behavior to radically new contexts upon deployment \citep{shi2024continual}. One important step towards this goal is to create systems that can induce rich representations of data that are seen in-context, and then flexibly deploy these representations to accomplish goals \citep{lampinen2024broader}. Recently, \citet{parkiclr} demonstrated that current LMs are indeed capable of inducing such representation from context (i.e., in-context representation learning). The present study investigates whether LMs can \textit{use} these representations to complete simple downstream tasks.

We first assess whether open-weights LMs can use in-context representations for next-token prediction, and then probe models using a novel task, \textbf{adaptive world modeling}. In both tasks, we find evidence that open-weights LMs struggle to deploy representations of novel semantics that are defined in-context, \textit{even if they encode these semantics in their latent representations}. Furthermore, we assess closed-source, state-of-the-art reasoning models on the adaptive world modeling task, and demonstrate that even the most performant LMs cannot reliably leverage novel patterns presented in-context. Overall, this work seeks to inspire novel methods for encouraging models to not only encode information presented in-context, but to do so in a manner that supports flexible deployment of this information.\footnote{Code released \href{https://github.com/mlepori1/in_context_world_models/tree/main}{here}.}
\end{abstract}

\section{Introduction}
The advent of transformer-based language models (LMs) has radically changed how we think about the capabilities of deep neural networks. Tasks that once appeared out of reach, such as competition mathematics \citep{luong2025advanced}, multilingual question answering \citep{comanici2025gemini}, and expert-level scientific reasoning \citep{rein2024gpqa}, have all seen substantial progress as a direct result of LM advancements. These advanced capabilities have spurred an ongoing investigation into whether transformers trained on next-token prediction encode an internal model of the causal process that generated their training data (i.e., a ``world model''), an internal model that could support such inferences \citep{li2021implicit, toshniwal2022chess, li2023emergent, vafa2024evaluating, vafa2025has}. Artificial agents that are endowed with such internal world models may ultimately prove more self-consistent \citep{wong2025modeling}, generate fewer hallucinations \citep{wong2023word}, and generalize better beyond their training distribution \citep{wu2024reasoning, mccoy2024embers, gupta2025better}.

In order to ensure that models can operate in a dynamic world without retraining, we would prefer artificial agents that are capable of adapting to different environments upon deployment. LMs have demonstrated a form of this ability, exemplified by the emergence of in-context learning (ICL; \citealt{brown2020language}). ICL enables pretrained LMs to temporarily adopt new behaviors based on examples or descriptions provided in context, without updating the model's parameters \citep{davidson2025different}. A fully realized adaptable agent would be able to learn to represent rich, structured representations that are defined entirely in-context \textit{and} flexibly deploy\footnote{We consider a representation to be \textbf{deployable} for a particular context if it plays a causal role in responding to that context, otherwise (e.g., if the representation is decodable, but not used; \citet{decharms2000neural}) we consider the representation to be \textbf{inert} with respect to that context. A representation is \textbf{flexibly deployable} if it is deployable for contexts distinct from the one that gave rise to it (e.g., downstream tasks).} those representations to solve downstream tasks \citep{lampinen2024broader}. In other words, such an agent would be able to construct an \textit{in-context world model}.



\begin{figure*}[h]
    \centering
    \includegraphics[width=.9\linewidth]{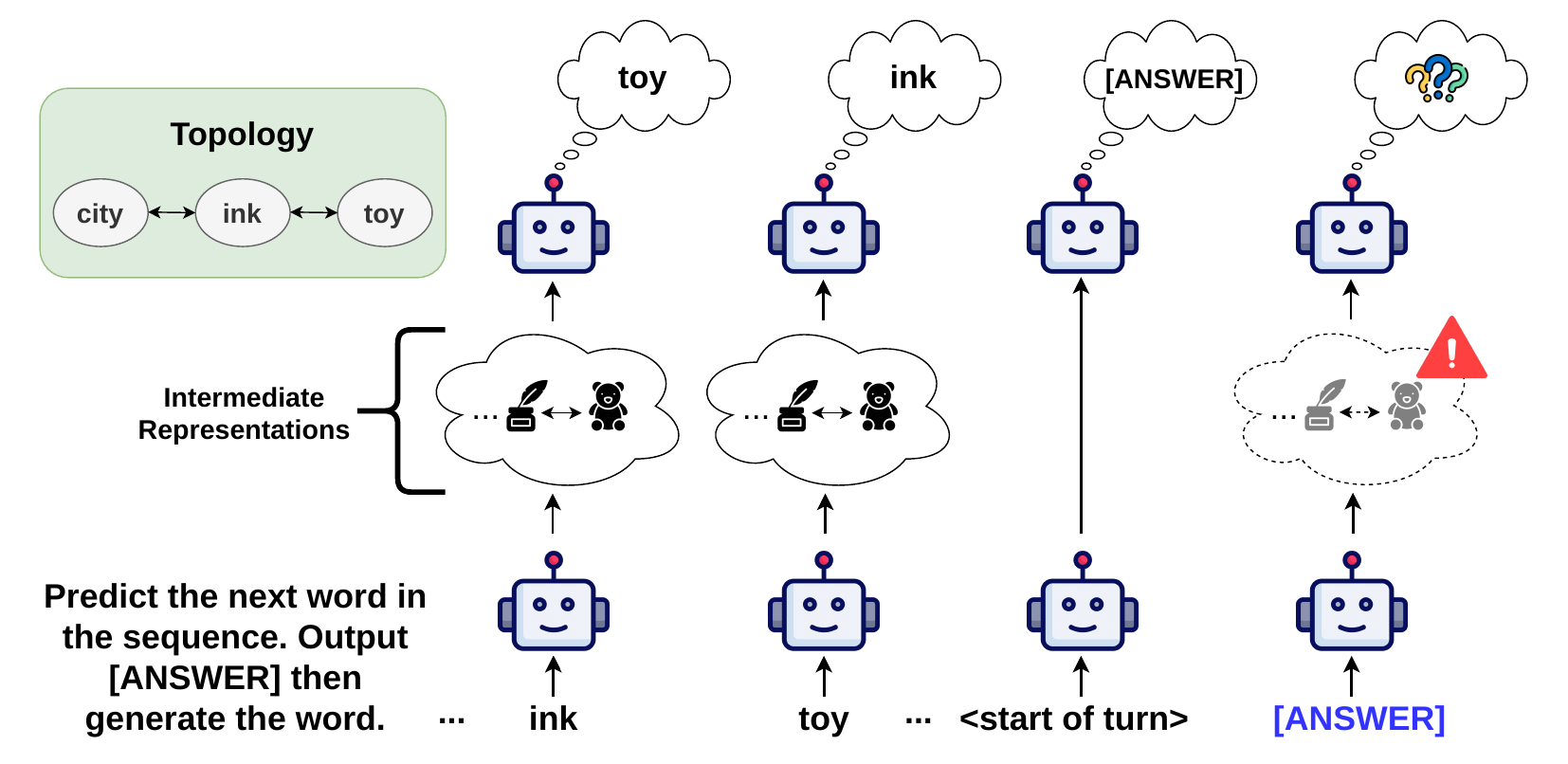}
    \caption{An illustration of a structured representation that is learned in context, but is not flexibly deployable. In the next-token prediction task, we present models with sequences of arbitrary  words generated by a random walk over a topology (in green), and ask the model to generate an [ANSWER] token, followed by the next token. The prompt tokens are represented in black, and the model-generated tokens are in blue. Given such a sequence, a language model may encode a representation of that topology within some of its hidden states after some number of layers. However, the model may be unable to properly deploy this representation later on when it is necessary for generating a response. In this work, we find that representations that are learned in context are oftentimes not flexibly deployable.}
    \label{fig:banner}
\end{figure*}

\begin{figure*}[h]
    \centering
    \includegraphics[width=.6\linewidth]{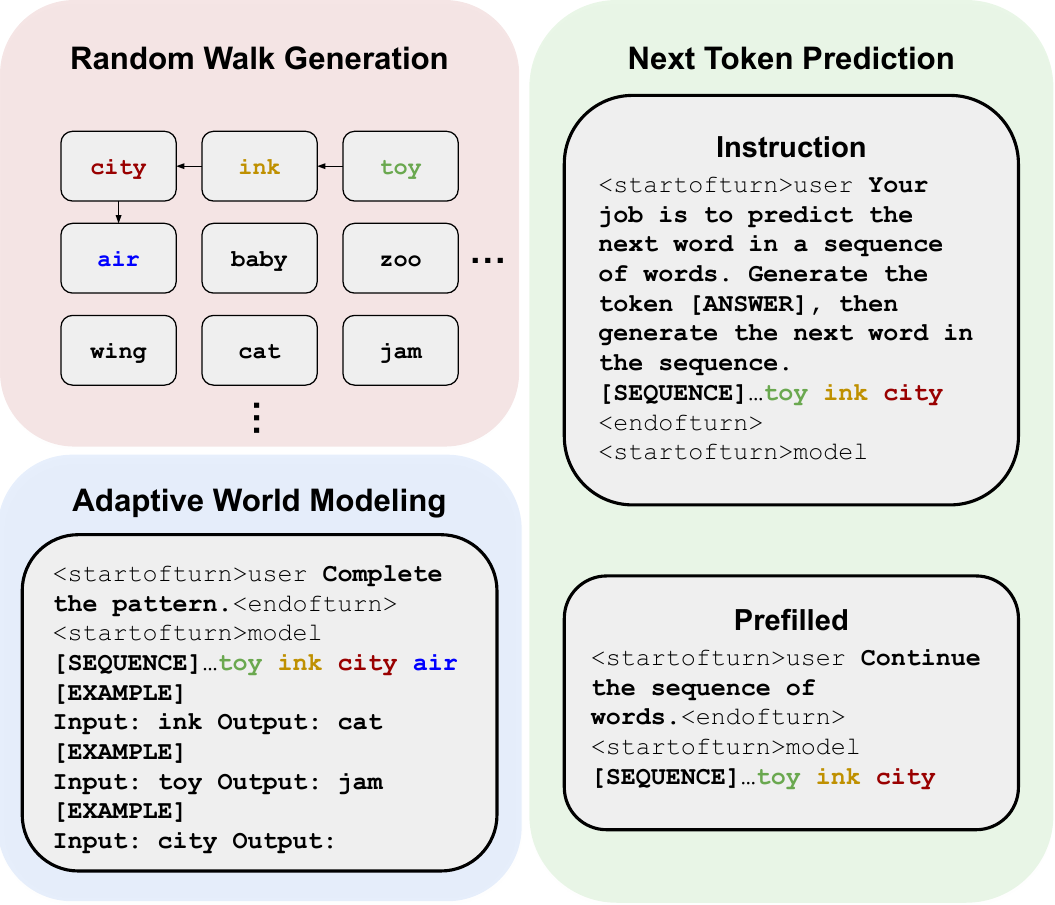}
    \caption{\textbf{(Top Left)} Example of an N-by-N state space topology used to generate a random walk. \textbf{(Right)} Examples of next token prediction prompts in the \textbf{Instruction} or \textbf{Prefilled} condition. Prompt formatting tokens are not bolded for readability. In the Instruction condition, models need to deploy in-context representations that are formed during the random walk (colored tokens) after an interval of several tokens in order to predict a valid next token. In the Prefilled Condition, in-context representations are deployed immediately. \textbf{(Bottom Left)} Example of an adaptive world modeling prompt, which consists of a random walk (colored tokens), followed by few-shot examples defining a rule that maps states at one position to states at another. Here the rule maps states $s_{i,j}$ to $s_{i+2, j}$ (a ``two-step rule''). Here, \textit{city} would get mapped to \textit{wing}.}
    \label{fig:Task}
\end{figure*}


A recent study explicitly investigates whether LMs can learn the kinds of structured internal representations required by an in-context world model. \citet{parkiclr} present LMs with a random walk over a state space with a latent topology, where each state is denoted by an arbitrary token (See Figures~\ref{fig:banner} and~\ref{fig:Task} for examples of these topologies). They demonstrate that LM token representations come to reflect the topology of that state space, moving the representations away from their default, uncontextualized semantics. Thus, LMs can in-context learn representations that encode completely novel semantics, a finding which has been replicated \citep{lubana2025priors} and used as inspiration for practical LM interventions \citep{yona2025context}. Here, we ask: are these in-context representations flexibly deployable when the model is faced with novel tasks, or are they inert in these conditions \citep{vafa2024evaluating, li2025just}? See Figure~\ref{fig:banner} for an illustration.

We investigate this question using prompts which crucially have two components: one which defines the in-context semantics (colored words in Figure~\ref{fig:Task}) and a separate component which requires the model to refer to these semantics to complete a task. We focus on two simple, synthetic tasks: (i) a next-word prediction task (Figure~\ref{fig:Task} Right), and (ii) a novel task, \textbf{Adaptive World Modeling} (Figure~\ref{fig:Task} Bottom Left). We find that in-context representations present in the hidden states of an LM are largely inert.

However, LMs may represent structures defined in context without encoding them in the geometry of their hidden states. We investigate one particularly salient alternative, exemplified by  state-of-the-art closed-source LMs. These so-called ``reasoning models'' are trained to externalize their intermediate reasoning steps into long chains of thought (CoTs; \citealt{wei2022chain, nyeshow, jaech2024openai}). Crucially, they appear to process information in a somewhat different manner than their base LM counterparts \citep{bogdan2025thought}, with CoT techniques enabling them to dynamically reason about a breadth of possible approaches to a given task \citep{DeepSeekAI2025DeepSeekR1IR, yao2023tree}, and even to behave as if they had constructed a model of the scenarios underlying a task \citep{coda2024cogbench}.  We investigate the possibility that these externalized reasoning chains can compensate for otherwise-inert hidden representations, enabling the creation of in-context world models. We find evidence that frontier reasoning models encode novel, in-context semantics in a manner that sometimes facilitates their use when solving downstream tasks, but they do so imperfectly, leaving substantial room for improvement.

In summary, our contributions are as follows:
\begin{enumerate}
    \item We introduce a new task, \textbf{adaptive world modeling}, that is designed to probe whether LMs can flexibly deploy novel, in-context semantics to solve simple tasks (Section~\ref{Sec:Exp2}).
    \item We find that open-weights models struggle to use in-context semantics to solve either next-token prediction tasks (Section~\ref{Sec:Exp1}) or adaptive world modeling tasks (Section~\ref{Sec:Exp2}), \textit{even when their latent representations encode these semantics.}
    \item We find that frontier reasoning models achieve limited success on adaptive world modeling tasks, suggesting that their use of long externalized reasoning chains may partially enable the use of in-context semantics (Section~\ref{Sec:Exp3}).
\end{enumerate}

\section{Preliminaries: In-Context Representation Learning}
\label{Sec:Preliminaries}

\paragraph{Graph Tracing Task}  We study LMs' in-context representation learning capabilities using the graph tracing task introduced by \citet{parkiclr}. The task is to predict the next word in a sequence of common, semantically unrelated one-token words. The sequence is generated by a random walk over a latent state space, with each state denoted by a unique word (See Figure~\ref{fig:Task}). In our experiments, these state spaces are organized into one of four topologies, defined by their dimensionality (2D grids or 1D lines) and total number of states (16 or 25). Specifically, we study 4-by-4 grids, 5-by-5 grids, 16-state lines, or 25-state lines, with many random assignments of words to states. As in prior work, we present a random walk over 2D topologies, and present a sequence of randomly selected pairs of adjacent states for 1D topologies. We measure how the representations of the words change as a function of context length. Using the metrics described below, we replicate \citet{parkiclr}'s main result: latent representations come to recapitulate the geometry of the underlying state space after sufficient context.

\paragraph{Metrics} Given a set of LM hidden representations corresponding to every token in the state space, we wish to characterize the extent to which these hidden states reflect the topology of the state space. Following prior work, we preprocess the representations by taking the mean of the hidden representations of all instances of each word in the state space at a given layer $l$ over a sliding window of 50 tokens in the random walk (or 50 adjacencies for 1D topologies). This gives us an average representation for each word at that layer, $H^l(\mathcal{T})$. We characterize in-context representation learning using two metrics:

    \textbf{Dirichlet Energy (DE):} This metric measures the distance between representations whose corresponding words are adjacent in the underlying state space. The core idea is that in-context representation learning should push adjacent words to be similar to one another. DE is defined as 

    \begin{equation}
                E_G(H^l(\mathcal{T})) = \sum_{i,j}A_{i,j}||h_i^l-h_j^l||^2
    \end{equation}
    where $E_G$ is the DE given state space topology $G$, $A_{i,j}$ is an indicator variable that denotes whether tokens $i$ and $j$ are adjacent in the state space, and $h_i^l$ refers to a particular hidden state of token $i$ at layer $l$. We normalize this metric by the total energy of the state space: $\sum_{i,j}||h_i^l-h_j^l||^2$.

    \textbf{Distance Correlation (DC):} This metric measures the correlation between distances in representation space and distances in the underlying state space. States that are far apart in the state space should correspond to representations that are far apart. Formally, DC is defined as
    \begin{equation}
        D_C(H^l(\mathcal{T})) = Corr(D_R^l, D_G)
    \end{equation}
    where
    \begin{equation}
                D_{R_{i,j}}^l = ||h^l_i-h^l_j||_2 \quad \text{and}\quad D_{G_{i,j}} =  ||G_i-G_j||_1
    \end{equation}
  
         $G_i$ refers to the coordinates of the token $i$ in the underlying state space. In other words, DC is the correlation between the Euclidean distance of the representations of particular tokens and the Manhattan distance between the states that these tokens correspond to. 
         
\paragraph{Replication}

\begin{figure}[h!]
    \centering
    \centering
    \includegraphics[width=\linewidth]{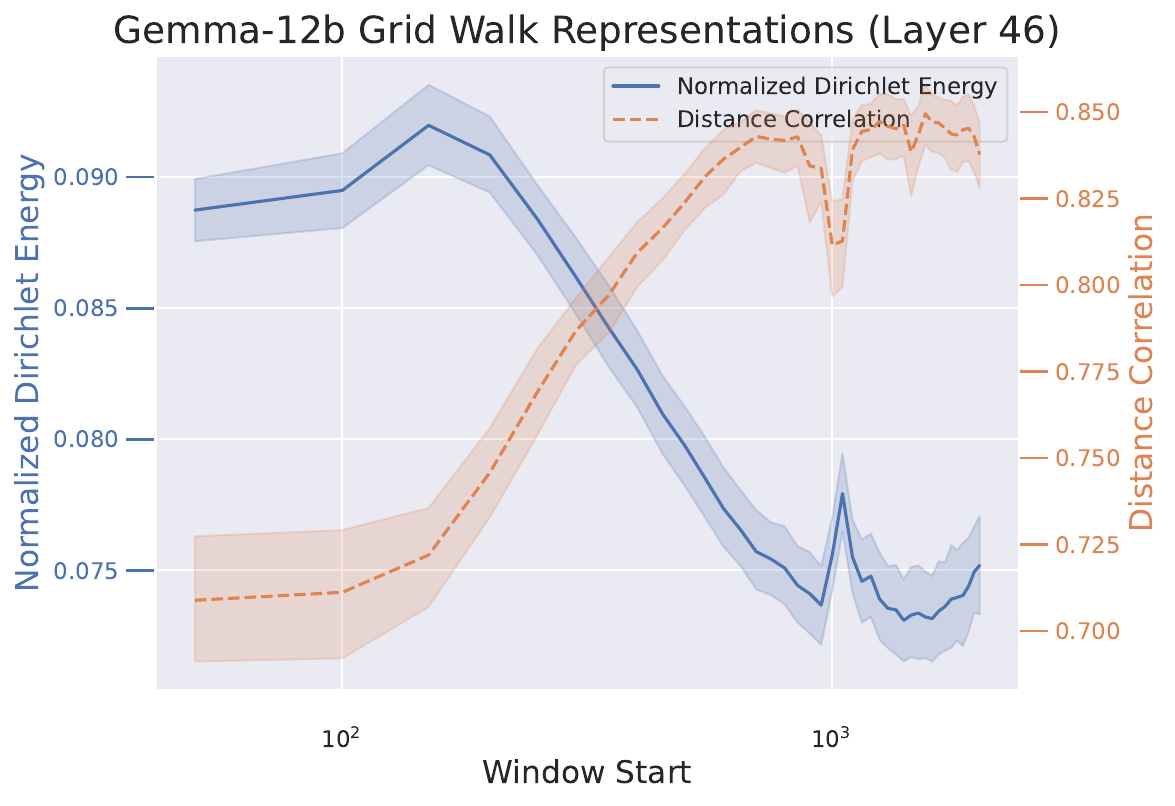}
    
    \caption{Example of in-context representation learning over a 5-by-5 grid topology. \textcolor{blue}{Dirichlet Energy} \textit{decreasing} and \textcolor{orange}{Distance Correlation} \textit{increasing} indicates better in-context representation learning.} 
    \label{fig:Replication}
\end{figure}
We study four capable instruction-tuned open-weights models, \texttt{gemma-3-\{4, 12, 27\}b-it} \citep{team2025gemma} and \texttt{OLMo-2-13b} \citep{olmo20242}. 
In general, we find that models can successfully learn representations in-context --- as context length increases, representations increasingly conform to the underlying state space topology. See Figure~\ref{fig:Replication} for results taken over 50 different token assignments on a 5-by-5 grid for an arbitrary late layer in \texttt{gemma-3-12b-it}.  We note some variation in the dynamics of in-context representation learning for some model-topology combinations, including some combinations that exhibit in-context representation learning early and then  degrade (See Appendix~\ref{App:Further_ICLR}).
 
\section{Experiment 1: In-Context Representation Learning Does Not Imply Robust Next-Token Prediction}
\label{Sec:Exp1}


To assess whether in-context representations are flexibly deployable, we take inspiration from prior work that identified a gap in next-word prediction performance when LMs are forced to delay their predictions \citep{huauxiliary}. While this work studied LM behavior on naturalistic sentences, we propose to extend their results into our synthetic graph tracing setting where we can quantify whether the representations encode the semantics required to perform next-word prediction. This allows us to disentangle failures of encoding semantics from failures of deploying these semantics.  

Specifically, we vary whether the random walk is presented in a user message (Instruction Condition), or whether it is presented in a prefilled model response (Prefilled Condition; See Figure~\ref{fig:Task} Right). In the Instruction Condition, the model is instructed to predict the next word in its response, which requires the model to delay using the in-context learned representations (due to the presence of intermediate special tokens). In contrast, the Prefilled Condition requires that the model uses the in-context representations immediately to perform next-word prediction. A response is considered ``correct'' if it is in the set of valid next tokens, given the preceding sequence and underlying topology.



\begin{figure}[]
    \centering
    \centering
    \includegraphics[width=\linewidth]{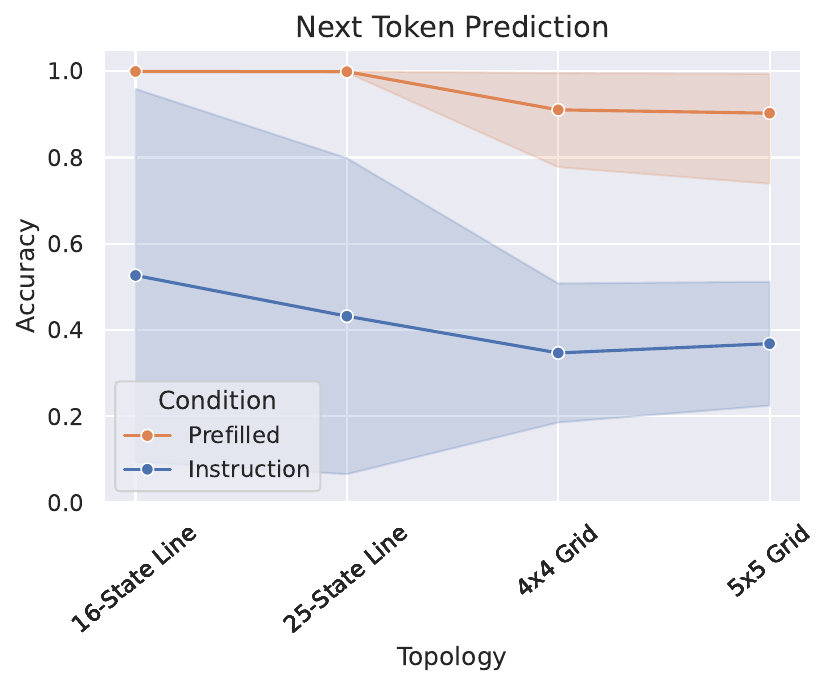}
    
    \caption{Next-token prediction results for all open-weights models over all topologies. Models struggle when the random walk is present in the user prompt, rather than in a prefilled model response. This indicates that the representations learned in-context during the random walk/random adjacencies are not easily deployed when they need to be used at a later time.}
    \label{fig:Exp1}
\end{figure}

\paragraph{Results} For each model and state space topology used in Section~\ref{Sec:Preliminaries}, we identify the context length the maximizes the average distance correlation achieved in the last and third-to-last layers (See Appendix~\ref{App:Ctx_Lengths} for the specific values). We run the next-word prediction analysis with 1000 different word assignments across all models and all state space topologies. Our results point to a  disparity between conditions: models reliably predict a valid next-word in the random walk when the random walk is prefilled (replicating prior work \citep{parkcompetition, lubana2025priors}), but struggle when the random walk is presented in the user prompt (See Figure~\ref{fig:Exp1}).

\begin{figure}
    \centering
    \includegraphics[width=\linewidth]{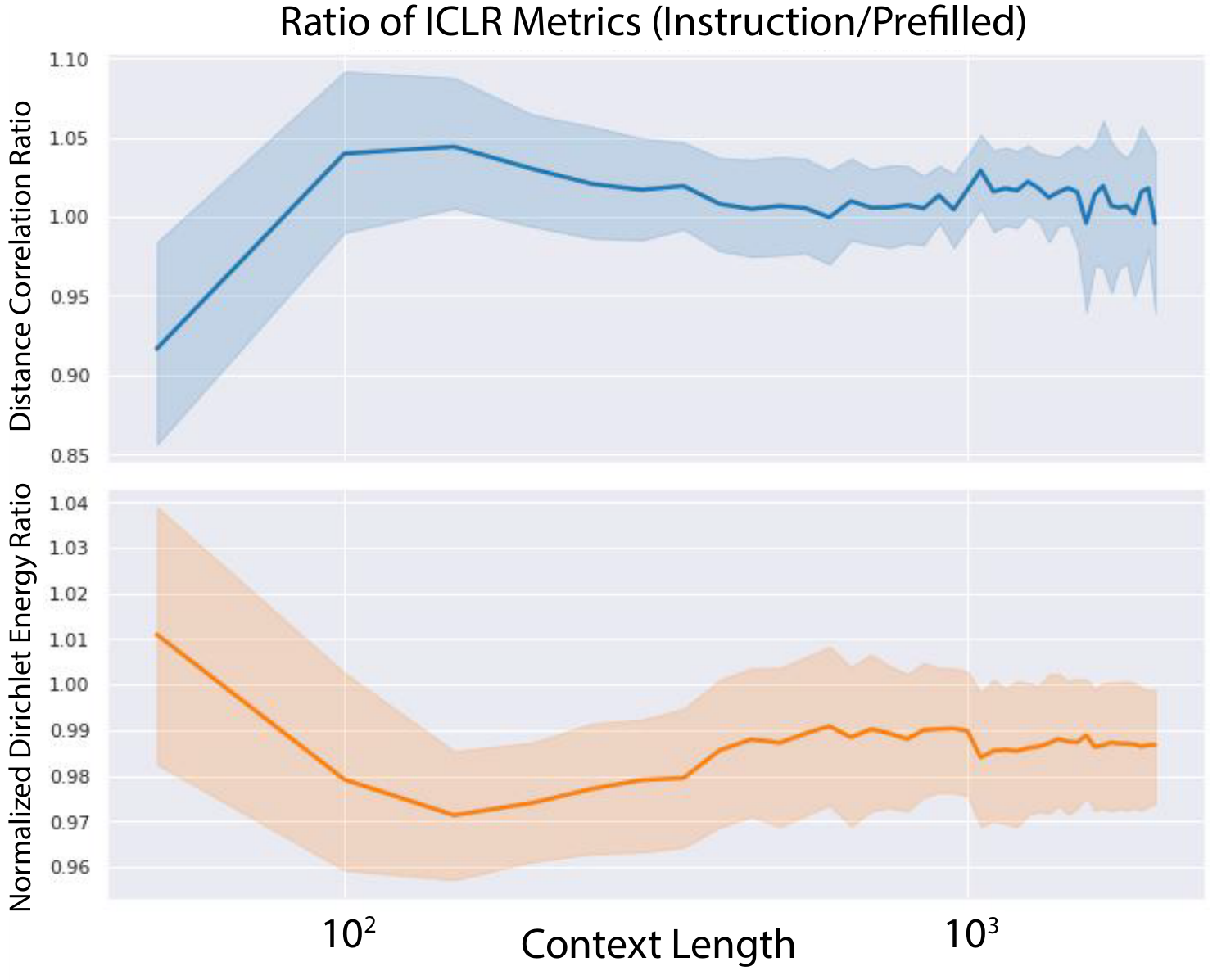}
    \caption{Ratio of in-context representation learning metrics between prompt format conditions over context. There does not appear to be a substantial difference between conditions (i.e., the ratio of metrics is near 1). If there is a small difference, it appears to be in-favor of instruction prompt representations, rather than prefilled prompt representations.}
    \label{fig:ICLR_Ratio}
\end{figure}

\paragraph{Analysis} 
One potential explanation for the performance disparity is that in-context representation learning simply yields worse representations in the Instruction Condition than the Prefilled Condition. However, we demonstrate that this is not the case in Figure~\ref{fig:ICLR_Ratio}. We compute the ratio of both in-context representation learning metrics over all models and topologies across all context windows (comprising 50 states in a random walk or 50 sampled adjacencies, depending on the dimensionality of the underlying topology). We find that in-context representation learning occurs in both conditions approximately equally. In fact, the small distinction that exists suggests higher fidelity representations in the Instruction Condition, thus ruling out the possibility that the Instruction Condition results in worse representations than the Prefilled Condition.

One might also worry that the context lengths we selected are insufficient for accurate next-token prediction. Perhaps in-context representations are \textit{initially} inert, but then become flexibly deployable after additional context. Indeed \citet{parkiclr} note that accurate next-token prediction sometimes requires a longer context length than in-context representation learning. We investigate this possibility in Appendix \ref{App:Long_NTP}, and find that models still struggle in the Instruction Condition with extended context lengths. Taken together, these results suggest that representations that are learned in-context are largely inert.

\section{Experiment 2: In-Context Representation Learning Does Not Imply Adaptive World Modeling}
\label{Sec:Exp2}
In Section~\ref{Sec:Exp1}, we identified a subtle divergence between in-context representation learning and next-token prediction performance, suggesting that in-context representations are not flexibly deployable. Here, we define a task, \textbf{Adaptive World Modeling (AWM)} to further probe this.

\paragraph{Task}
AWM prompts consists of two components. The first is the graph tracing task (defined in Section~\ref{Sec:Preliminaries}). Models must infer the latent state space topology from a random walk over that topology. The second component is a few-shot learning task, where elements at one position in the topology get mapped to elements at another position according to a simple rule (e.g., state $s_{i,j} \rightarrow s_{i+2, j}$; See Figure~\ref{fig:Task}). Because completely arbitrary words are used to denote each state, LMs \textit{must} deploy the structure that they have encoded in-context to solve the task.

We investigate the same set of topologies presented in Section~\ref{Sec:Preliminaries}, and investigate several different rules that govern  the few-shot learning task. For all topologies, we investigate a ``1-step down'' rule (where the input-output transitions are attested in the random walk) and ``2-step down'' rule (where they are not; see Figure~\ref{fig:Task}). For the 5-by-5 grid, we additionally investigate a ``3-step rule'', where states at position $(i,j)$ get mapped to states at position $(i+2, j+1)$.

In most settings, we include 10 examples in the context and query the model with a held-out example. Because the 4-by-4 grid topology only contains 8 well-defined ``2-step down'' examples, we include 6 examples for that prompt. We present the prompts as prefilled model responses\footnote{We also tested \texttt{gemma-3-27b-it} on a version of the task where the instructions, random walk, and few-shot examples are presented in the user message, and achieve approximately the same poor performance.} and use the context lengths used in Section~\ref{Sec:Exp1}. We evaluate on 1000 token assignments for each condition.

\begin{figure}
    \centering
    \includegraphics[width=\linewidth]{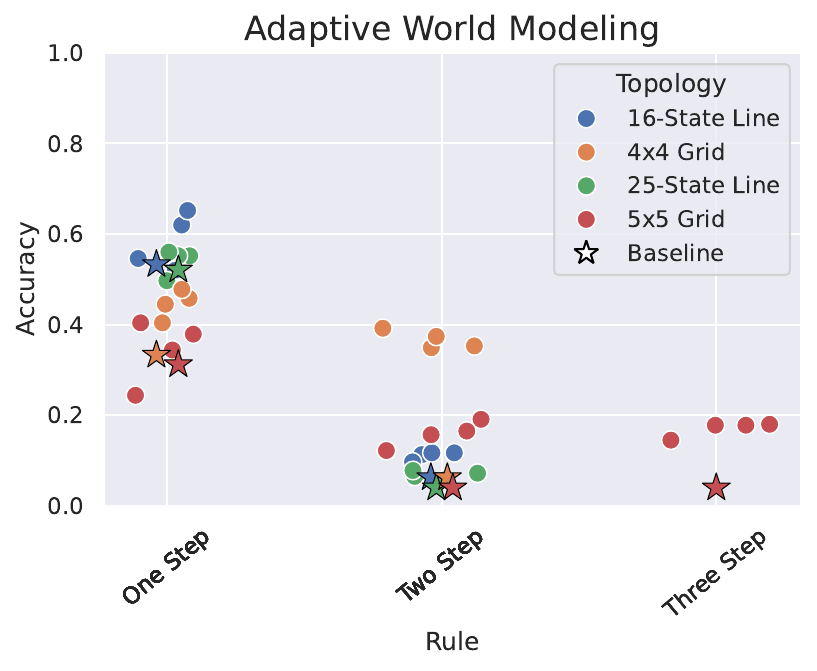}
    \caption{Adaptive World Modeling results. Across all task configurations, various open-weights LMs (each represented by a separate dot) struggle, despite encoding the underlying grid topology in their latent representations. Note: A naive strategy of randomly selecting an attested adjacency would achieve approximately 50\% (for one-dimensional) and 30\% (for two-dimensional) accuracy for the ``one-step'' rules.}
    \label{fig:Exp2_ICWM}
\end{figure}

\begin{figure}[h!]
    \centering
    \begin{minipage}{\columnwidth} 
        \centering
        \includegraphics[width=\textwidth]{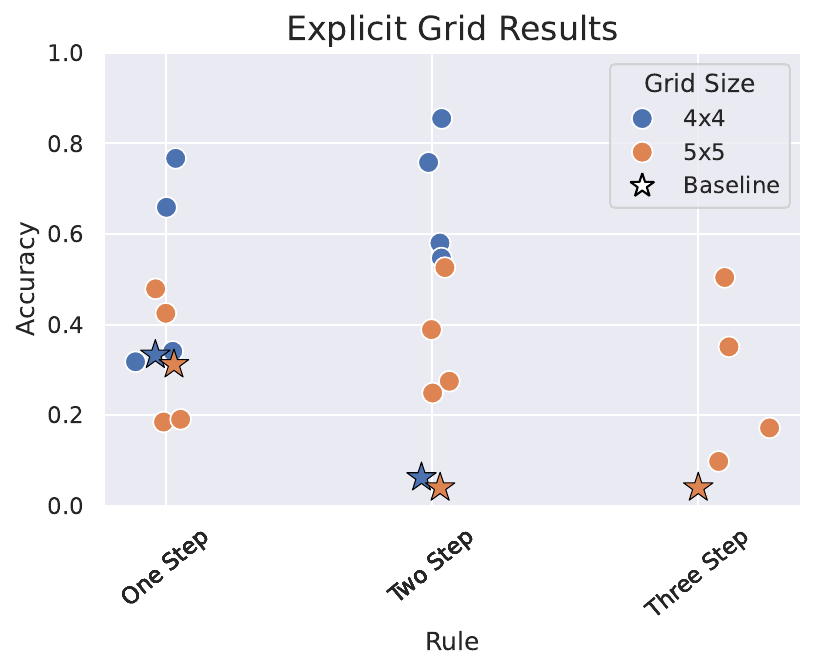}
    \end{minipage}\par 
    
    \begin{minipage}{\columnwidth}
        \centering
        \includegraphics[width=\textwidth]{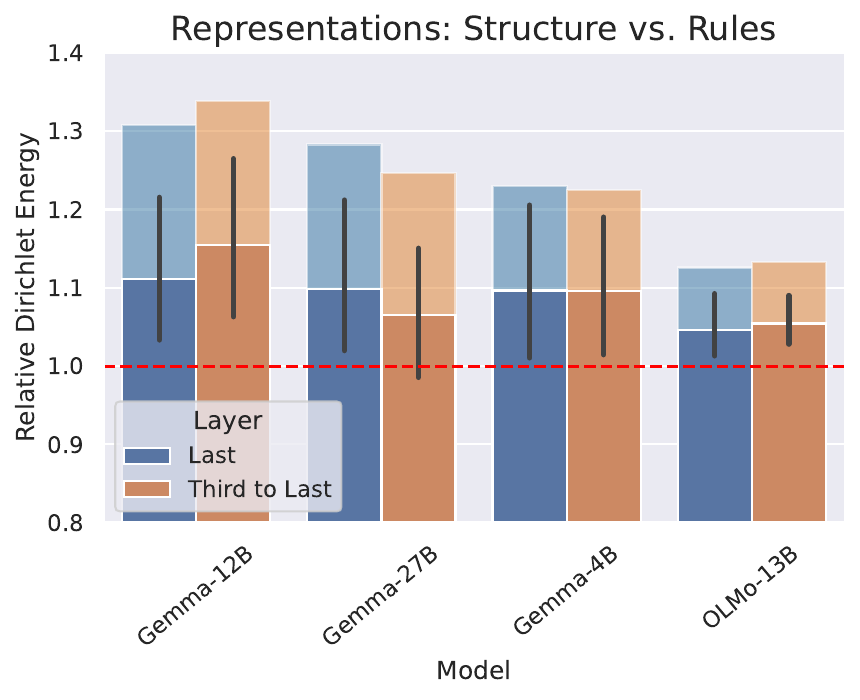}
    \end{minipage}
    \caption{(Top) Open-weights LM results on few-shot learning when the state space topology is explicitly presented in the prompt. LMs achieve variable performance, but it is typically much higher than in the AWM setting. (Bottom) Relative Dirichlet Energy between tokens in the random walk and in the few-shot examples (and between random walk tokens and uncontextualized tokens; light bars) over two layers. In-context representations encode the underlying topology with lower fidelity in the few-shot examples.}
\label{fig:Exp2_Analysis}
\end{figure}

\paragraph{Results}
 From Figure~\ref{fig:Exp2_ICWM}, we see that all open-weight LMs that we tested struggle with the AWM  task. LMs may seem to perform somewhat better at the one-step rules, but these rules allow for a trivial baseline strategy of randomly guessing a transition observed in the random walk, rather than truly inferring the underlying rule (see Appendix~\ref{App:Naive} for details). However, we do see that some topology-rule combinations permit nontrivial performance on the AWM task. In particular, models achieve approximately 40\% accuracy on the 4-by-4 grid topology for one-step and two-step rules.

 \paragraph{Analysis} An LM that has properly formed in-context representations corresponding to the underlying state space might struggle to solve an AWM task for one of two reasons: (i) the LM cannot properly induce the rule from the few-shot examples, even with perfect information about the topology, or (ii) the in-context learned representations are largely inert in the context of an AWM task, i.e., the LM cannot reliably use the topology information encoded in them when confronted with the few-shot examples. We investigate both of these alternatives. 

First, we investigate the model's few-shot learning ability. We provide an explicit description of the underlying state space for the 4-by-4 or 5-by-5 grid topologies in the prompt and then provide few-shot examples of the rule before querying the model with a held-out state. Specifically, we include statements of the form \texttt{Coordinates: 0 0 Item: city} to describe the grid topology. In Figure~\ref{fig:Exp2_Analysis} (Top), we see that LMs achieve substantially higher accuracy when given the explicit description than when they need to infer the topology in-context, with some models achieving $>$ 75\% accuracy on AWM tasks defined on the 4-by-4 grid with one-step and two-step rules. Thus, LMs' struggles with AWM cannot be fully attributed to their ability to learn the rule from examples.

Next, we investigate whether the representations learned in-context are flexibly deployed during the few-shot learning portion of the AWM task. In particular, we investigate whether the hidden representations of words present in the few-shot examples also encode the state space topology. If representations do not reflect the state space topology when words are presented in the few-shot learning examples, then this provides evidence that in-context representations are not being flexibly deployed to learn the rule.
We measure the ratio of the Dirichlet Energy (DE) of representations drawn from the random walk to the DE of the same tokens when they are present in the few-shot examples. If this ratio is greater than 1, then the representations present in the examples encode the underlying topology with lower fidelity than the representations in the random walk (see Appendix~\ref{App:Rel_Dirichlet} for details). For comparison, we also compute the ratio of the DE of uncontextualized tokens to the DE of random walk tokens. In Figure~\ref{fig:Exp2_Analysis} (Bottom), we find that representations in the few-shot learning examples are degraded relative to their counterparts in the random walk. This suggests that the in-context learned representations are not being flexibly deployed to solve the adaptive world modeling task.

One might be concerned that LMs simply do not infer that the few-shot examples are at all related to the random walk. We test a meta-learning variation of the task where examples of successful completions are prepended to the prompt, but find that models still perform poorly (See Appendix~\ref{App:metalearning}). 
Similarly to Section~\ref{Sec:Exp1}, we additionally investigate longer contexts in Appendix~\ref{App:Long_ICWM}, and find that this also does not improve performance.
Taken together, these results further support the claim that representations learned in-context are largely inert (i.e., unavailable for solving downstream tasks).

\section{Experiment 3: Investigating Reasoning Models}
\label{Sec:Exp3}

Finally, we investigate whether our analyses generalize to frontier models: \texttt{Gemini-Flash-2.5}, \texttt{Gemini-Pro-2.5} \citep{comanici2025gemini}, \texttt{GPT-5-mini}, and \texttt{GPT-5} \citep{gpt5}. These models are so-called ``reasoning models'', which generate extended, verbalized reasoning chains before providing an answer to a user query \cite{nyeshow}. This form of computation represents the state-of-the-art for modern LMs and has resulted in substantial improvements in tasks that require advanced reasoning \citep{comanici2025gemini, gpt5, DeepSeekAI2025DeepSeekR1IR}.

However, the representations and mechanisms supporting reasoning models remain even more opaque than the inner workings of standard LMs \citep{bogdan2025thought, macar2025thought}, with some arguing that reasoning models implement fundamentally new algorithms when using extended reasoning chains \citep{li2025system}. Can these advanced reasoning models offer a means of encoding in-context semantics and flexibly deploying them to solve downstream tasks? In this section, we investigate this question by testing frontier LMs on the AWM task.

\paragraph{Results}
Similarly to our open-weight model analyses in Appendices~\ref{App:Long_NTP} and \ref{App:Long_ICWM}, we provide models with 1500-token random walks or 1500 pairs of adjacent states for each topology, and use the same rules as described in Section~\ref{Sec:Exp2}. For \texttt{Gemini} models, we let the model generate up to 5000 tokens in its chain-of-thought before responding. We impose this ``thinking budget'' to prevent timeouts caused by extremely long thinking generations.
For \texttt{GPT-5} models, we enable the model to flexibly determine its thinking budget. We do not view this discrepancy as an issue, as our primary question is whether \textit{any} LMs can perform well on the adaptive world modeling task. Furthermore, initial experimentation revealed that imposing a 5000-token thinking budget did not substantially impact the performance of \texttt{Gemini} models. 

\begin{figure}[]
    \centering
    \includegraphics[width=\linewidth]{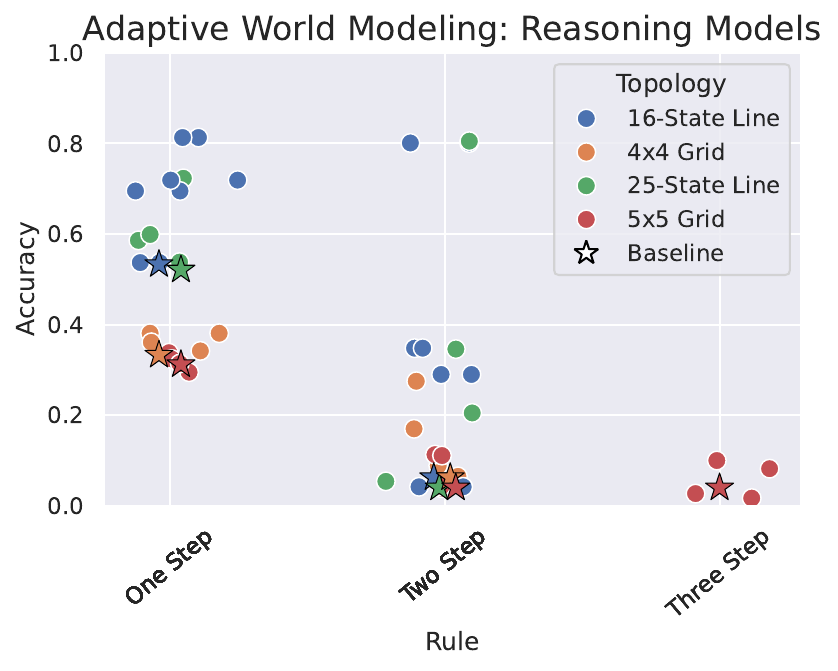}
    \caption{Frontier LM results on AWM. Frontier LMs achieve higher performance than open-weights models, especially for one-dimensional (line) topologies. However, performance drastically degrades when they are confronted with two-dimensional (grid) topologies.}
    \label{fig:Exp3_ICWM}
\end{figure}

Our results are presented in  Figure~\ref{fig:Exp3_ICWM}. We see that even these frontier reasoning LMs do not completely solve the AWM task, though many perform markedly better than their open-weights counterparts. In particular, some models achieve nontrivial accuracy on 1D topologies for one-step and two-step rules. However, accuracy entirely collapses when models are presented with grid topologies. 

\paragraph{Analyses}
Similarly to Section~\ref{Sec:Exp2}, we investigate two potential sources for LM errors on the AWM task: 1) Not being able to induce the rule from the few-shot learning examples (even with perfect knowledge of the underlying state space topology) and 2) Not being able to produce a representation of the underlying state space topology that is able to be deployed for downstream tasks. We investigate \texttt{Gemini-2.5-\{Pro, Flash\}}.

\begin{figure}[]
    \centering
    \includegraphics[width=.9\linewidth]{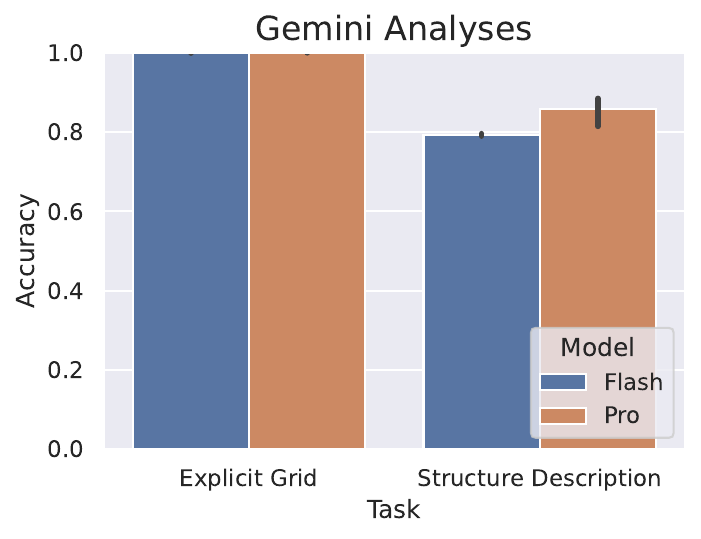}
    \caption{(Left) When presenting either \texttt{Gemini} model with a prompt that explicitly describes the underlying state space topology, the LM achieves ceiling accuracy on a few shot learning task. (Right) When prompting \texttt{Gemini} models to describe the structure of the latent state space that generated a random walk, models achieve fair, but not perfect, accuracy (even when assessed using a lenient autorater).}
\label{fig:Exp3_Analysis}
\end{figure}

First, we investigate whether frontier reasoning LMs can solve the few-shot learning task when they are presented with an explicit representation of the underlying state space. This analysis proceeds exactly as in Section~\ref{Sec:Exp2}. From Figure~\ref{fig:Exp3_Analysis}, we see that both models achieve ceiling performance in this setting. Thus, errors are \textit{not} caused by a failure to learn the rule from the examples.

Next, we investigate whether these models can induce a flexible representation of the underlying state space topology. Because we do not have access to the hidden representations of these frontier models, we investigate this question by directly asking the model to describe the structure of the state transition matrix underlying a random walk. We provide models with 1000 random walks over 4-by-4 and 5-by-5 grids. Succeeding at this requires the model to (i) have inferred the underlying topology in the first place and (ii) be able to deploy these encoded in-context semantics to generate a natural language description of them. To assess whether the model successfully describes the state space, we employ a permissive autorating strategy. Specifically, we provide each LM response in a new prompt to \texttt{Gemini-2.5-Pro}, and ask the model to respond ``yes'' if the response contains a reference to a 4-by-4 (or 5-by-5) grid, and otherwise respond ``no''. Note that the original LM description does not need to actually infer the correct word-state mapping to count as correct. Nonetheless, we find that both \texttt{Pro} and \texttt{Flash} models achieve good, but not perfect performance (See Figure~\ref{fig:Exp3_Analysis}). By construction, the LMs' rate of achieving a perfect understanding of the underlying state space is even lower. Thus, it appears that errors are driven primarily by the LMs' encoding and deployment of a flexible representation of the state space topology.

In Appendix~\ref{App:Frontier_NTP}, we present frontier model results from the next-token prediction task. We find that most of these models models perform reasonably well. In Appendix~\ref{App:Ablating_Thinking} we investigate the role of reasoning chains on task performance for both next-token prediction and adaptive world modeling by ablating the reasoning chains in \texttt{Gemini} models. We find mixed results, with performance on both tasks sometimes substantially \textit{increasing} for \texttt{Flash} with reasoning chains ablated, and performance on both tasks systematically dropping for \texttt{Pro}. Finally, in Appendix~\ref{App:Hints} we find that additional guidance on the underlying state space topology improves model performance, providing further evidence that representations of the underlying topology enable reasoning models to succeed at AWM.

Taken together, these results suggest that externalized reasoning chains may provide an alternative pathway for models to encode and deploy representations of semantics that are defined in-context. However, the circumstances in which these representations may be deployed is somewhat limited, leaving much room for improvement in the development of adaptable artificial intelligence systems.

\section{Discussion}

\paragraph{Related Work}
The present work is situated at the intersection of two active research directions: (i) characterizing the internal world models implemented by language models and (ii) characterizing the capabilities and limitations of in-context learning.
Several works have evaluated world models by probing their internal representations \citep{li2023emergent, gurneelanguage, li2021implicit} or generalization behavior \citep{vafa2024evaluating, vafa2025has}. Many of these works rely on a model's ability to track state throughout context, a precursor to the in-context representation learning under study in the present work. Finally, several theoretical works have attempted to define a world model \citep{andreasWorldModels, li2025does}. One common theme is that a world model is canonically \textit{not} just a look-up table, and instead affords other manipulations. In the present context, this implies that an inert set of representations learned in-context does not constitute an in-context world model.

With respect to in-context learning, much research has focused on characterizing ICL's algorithmic properties \citep{akyureklearning, von2023transformers}, mechanistic implementation \cite{olsson2022context, toddfunction, hendel2023context}, and the conditions for its emergence \citep{chan2022data, parkcompetition, wurgaft2025context}. However, the study of in-context representation learning is still in its infancy \citep{parkiclr, lubana2025priors, yona2025context}.

\paragraph{Conclusion}
In this work, we investigated whether LMs could flexibly deploy representations of in-context defined semantics to solve downstream tasks. Across several tasks and models, our results suggest that LMs cannot reliably do so. Crucially, even when LMs encode these semantics in their latent representations, these representations are often \textit{inert}, or unable to be used for either next-token prediction or adaptive world modeling. Though frontier models seem somewhat more adaptable than their non-reasoning counterparts, the creation of truly adaptable artificial agents requires training regimens and architectures that support flexible deployment of representations learned in-context. 

\section{Limitations}
This work has several limitations. Notably, we reveal that in-context semantics encoded in hidden representations may not be flexibly deployable, but do not offer a solution. We speculate that either targeted training regimens or mechanistic interventions that make these semantics available to earlier model layers (similar to those found in \citet{biran2024hopping} and \citet{lepori2025racing}) can partially address this issue.

Furthermore, it is difficult to understand the relationship between the externalized chains-of-thought produced by reasoning models and their hidden token representations. While we provide evidence that models that employ chain-of-thought seem to partially overcome the limitations of in-context learned representations, we do not yet understand the mechanisms by which this occurs. Future work might focus on clarifying this relationship.

\section*{Acknowledgments}
The authors would like to thank Andrew Lampinen, Martin Wattenberg, Asma Ghandeharioun, Effie Li, Eghbal Hosseini, Yasaman Bahri, Julian Zimmert, and the members of the Computation and Psycholinguistics Lab at NYU for their helpful feedback. The authors would also like to thank Rachel Goepner, for proofreading the manuscript. 

This material is based upon work supported by the National Science Foundation Graduate Research Fellowship under Grant No. 2439559. Any opinions, findings, and conclusions or recommendations expressed in this material are those of the author(s) and do not necessarily reflect the views of the National Science Foundation.

\bibliography{custom}

\appendix

\section{Further Examples of In-Context Representation Learning}
\label{App:Further_ICLR}
\begin{figure}
    \centering
    \includegraphics[width=\linewidth]{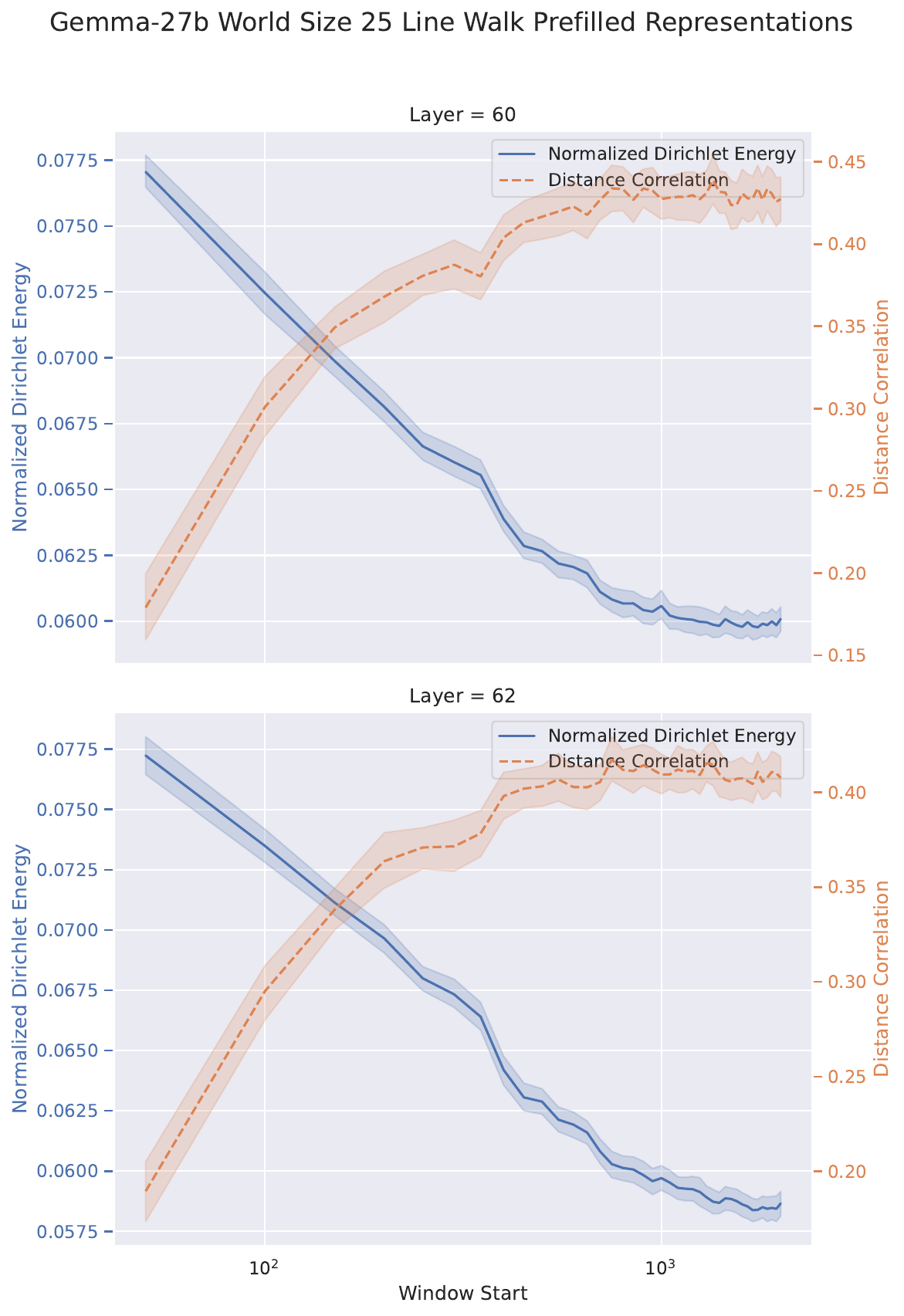}
    \caption{\texttt{Gemma-3-27b-it} in-context representation learning on a 25-length line topology.}
    \label{fig:App_Ex1}
\end{figure}

\begin{figure}
    \centering
    \includegraphics[width=\linewidth]{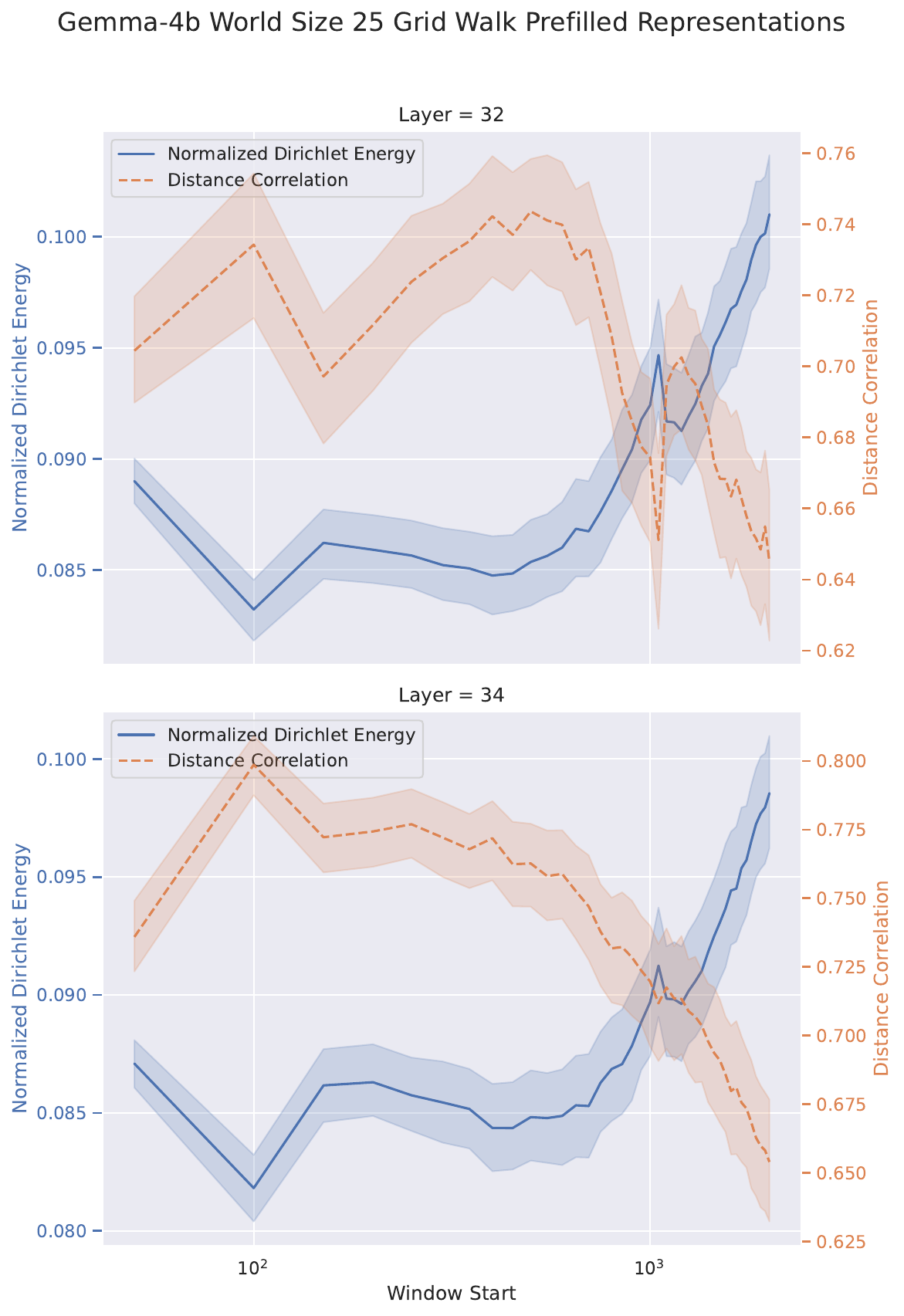}
    \caption{\texttt{Gemma-3-4b-it} in-context representation learning on a 5-by-5 grid  topology. In this case, in-context representation learning is less well-behaved, though the distance correlation metric still increases over context.}
    \label{fig:App_Ex2}
\end{figure}

\begin{figure}
    \centering
    \includegraphics[width=\linewidth]{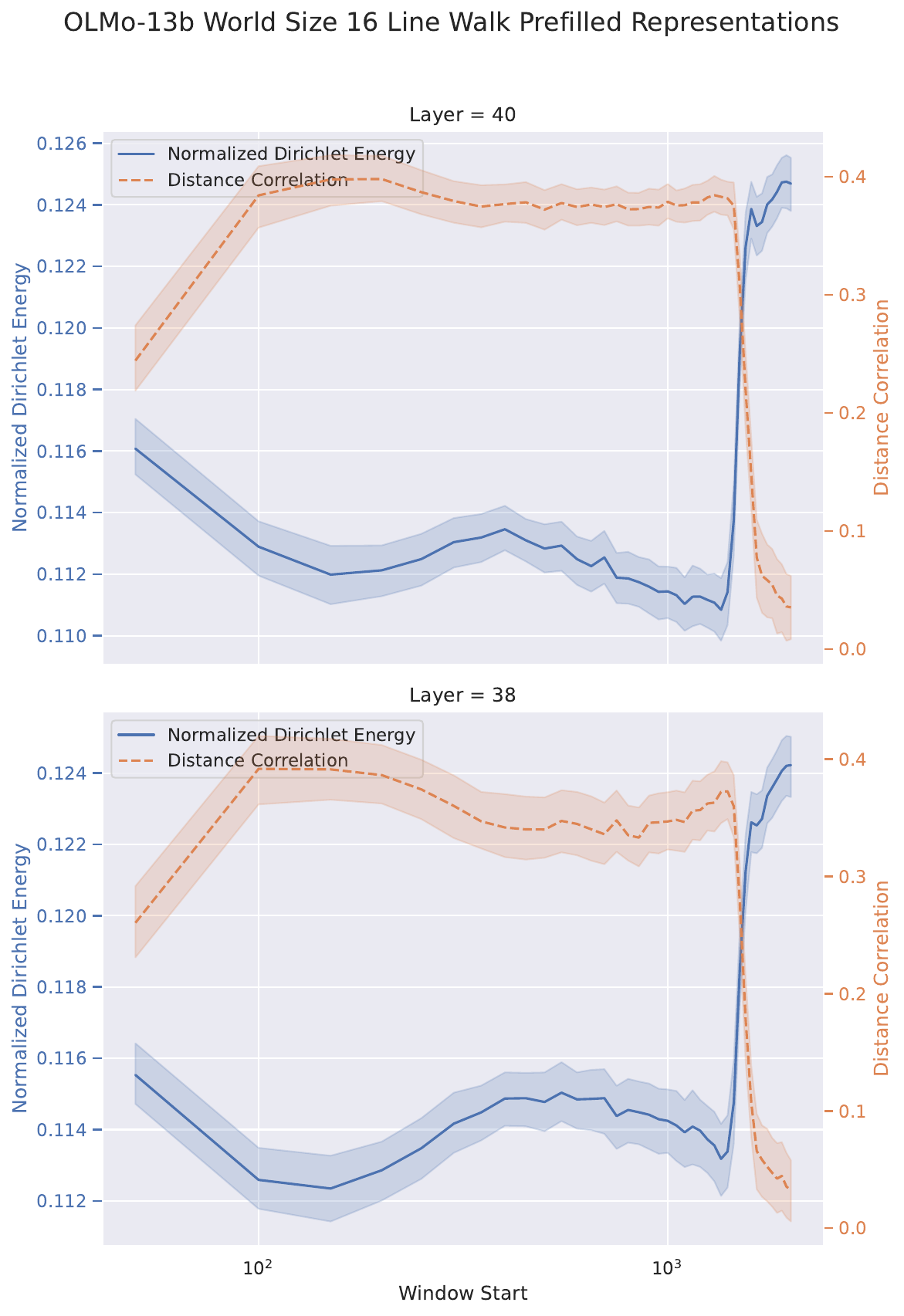}
    \caption{\texttt{OLMo-2-13b} in-context representation learning on a 16-length line  topology. In this case, in-context representation learning occurs rapidly. This collapses much alter in the context.}
    \label{fig:App_Ex3}
\end{figure}

In this section, we present three more examples of in-context representation learning over different topologies and different models. See Figures~\ref{fig:App_Ex1}, \ref{fig:App_Ex2}, and \ref{fig:App_Ex3} for a particularly diverse set of examples.

\section{Optimal Context Lengths Per Model and Topology}
\label{App:Ctx_Lengths}

Table~\ref{tab:context_lengths} reports the optimal contexts lengths from the analysis of in-context representation learning described in Section~\ref{Sec:Exp1}.

\begin{table}[]
    \centering
    \begin{tabular}{c|c|c|c}
         Model & Prompt & Topology & Ctx. Len \\
         \hline
         Gemma-27b & Prefill & 16-Line & 1200 \\
         Gemma-27b & Prefill & 25-Line & 1350 \\
         Gemma-27b & Prefill & 16-Grid & 350 \\
         Gemma-27b & Prefill & 25-Grid & 650\\
         Gemma-12b & Prefill & 16-Line & 450 \\
         Gemma-12b & Prefill & 25-Line & 500 \\
         Gemma-12b & Prefill & 16-Grid & 850 \\
         Gemma-12b & Prefill & 25-Grid & 1550 \\
         Gemma-4b & Prefill & 16-Line & 1300 \\
         Gemma-4b & Prefill & 25-Line & 1500 \\
         Gemma-4b & Prefill & 16-Grid & 300 \\
         Gemma-4b & Prefill & 25-Grid & 100 \\
         OLMo-13b & Prefill & 16-Line & 150 \\
         OLMo-13b & Prefill & 25-Line & 200 \\
         OLMo-13b & Prefill & 16-Grid & 100 \\
         OLMo-13b & Prefill & 25-Grid & 800 \\
         \hline
         Gemma-27b & Instr. & 16-Line & 1150\\
         Gemma-27b & Instr. & 25-Line & 1600 \\
         Gemma-27b & Instr. & 16-Grid & 400\\
         Gemma-27b & Instr. & 25-Grid & 550\\
         Gemma-12b & Instr. & 16-Line & 300\\
         Gemma-12b & Instr. & 25-Line & 750\\
         Gemma-12b & Instr. & 16-Grid & 650\\
         Gemma-12b & Instr. & 25-Grid & 700\\
         Gemma-4b & Instr. & 16-Line & 1600\\
         Gemma-4b & Instr. & 25-Line & 1550\\
         Gemma-4b & Instr. & 16-Grid & 200\\
         Gemma-4b & Instr. & 25-Grid & 300\\
         OLMo-13b & Instr. & 16-Line & 1400\\
         OLMo-13b & Instr. & 25-Line & 1350\\
         OLMo-13b & Instr. & 16-Grid & 250\\
         OLMo-13b & Instr. & 25-Grid & 750\\
    \end{tabular}
    \caption{Optimal context lengths for in-context representation learning}
    \label{tab:context_lengths}
\end{table}

\section{Long Context Next Token Prediction}
\label{App:Long_NTP}
\begin{figure}
    \centering
    \includegraphics[width=\linewidth]{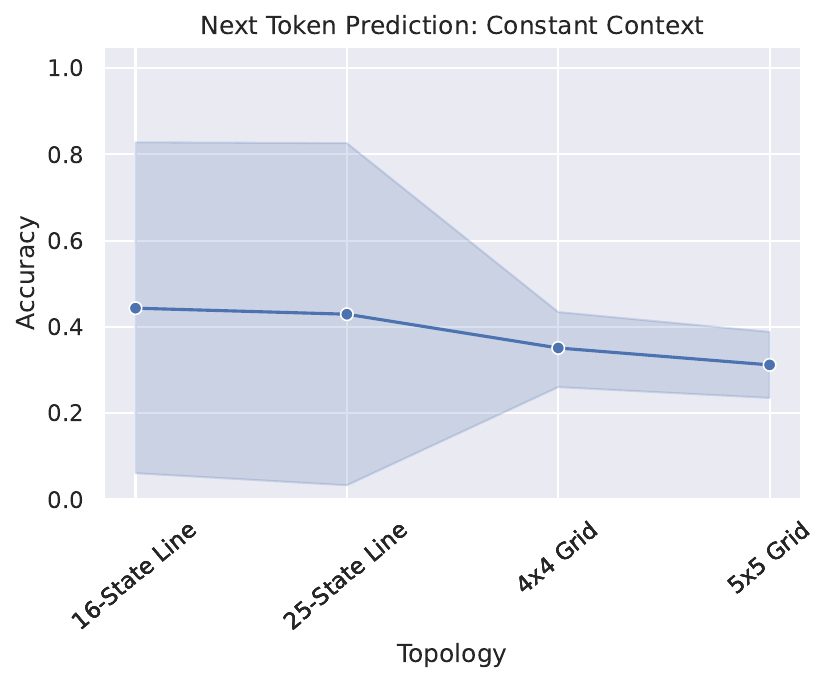}
    \caption{Next token prediction results with constant long context length.}
    \label{fig:ntp_cc}
\end{figure}

In this section, we present next-token prediction results for the next-token prediction task using the instruction prompt format, except we provide the all models with a consistent long context length of 1500. From Appendix~\ref{App:Ctx_Lengths}, we can see that this is longer than the previously analyzed context lengths for all but 3 model-topology combinations. From Figure~\ref{fig:ntp_cc}, we see that this extended context length does not help performance.

\section{Naive Solution for One Step Rules}
\label{App:Naive}

This section describes a simple heuristic solution to adaptive world modeling tasks with a one-step rule, and presents the accuracies that this solution achieves. First, note that the input/output pairs given by the one-step rules (i.e., those that map $S_i$ to $S_{i+1}$ or $S_{i, j}$ to $S_{i+1, j}$) are attested in the random walk. Thus, a simple solution is to guess one of the attested adjacent states to the input state. However, not every state corresponds to a valid input state for a one-step rule (e.g., with zero-indexing $S_{24}$ does not map to a valid state in a 25-state line, as $S_{25}$ does not exist). Additionally, not every state has the same number of adjacent states. In the case of a line, the states at either end only have a single adjacent state. Randomly sampling one of the attested adjacent states for each possible query example gives the accuracies presented in Table~\ref{tab:baselines}.

\begin{table}
\begin{tabular}{c|c|c}
   Dimension  & World Size & Baseline Acc. \\
   \hline
     1 & 16 & 53.3\%\\
     1 & 25 & 52.1\% \\
     2 & 16 & 33.3\%\\
     2 & 25 & 31.2\%\\

\end{tabular}
\caption{Naive baseline strategy accuracies.}
\label{tab:baselines}
\end{table}

\section{Measuring Relative Dirichlet Energy}
\label{App:Rel_Dirichlet}
In Section~\ref{Sec:Exp2}, we present an analysis of the relative Dirichlet Energy between tokens in the random walk and tokens in the few-shot examples in an adaptive world modeling task. In this section, we provide details on this analysis.

First, we run 50 new adaptive world modeling prompts for each model and topology. We collect activations from the last 50 tokens in a random walk (or 50 adjacencies in a line topology). As described in Section~\ref{Sec:Preliminaries}, we compute average token representations for each token. We repeat this process separately for tokens that appear in the portion of the prompt containing few-shot examples. Notably, these two sets of tokens will be subsets of the tokens that occupy the state space (this is guaranteed for the tokens in the few-shot learning portion of the prompt, as several tokens are held-out by design). To account for this, we take the intersection of tokens occurring in both sets. Then, we compute normalized Dirichlet Energy as described in Section~\ref{Sec:Preliminaries} for each of the sets independently, and report the ratio of these values. 

To provide context for these results, we extract uncontextualized representations of the tokens present in the intersection of the two sets described above (corresponding to tokens in the last window of the random walk, or in the few-shot examples). In this case, uncontextualized representations correspond to instances of these tokens that are at the very first position in the random walk. We compute and report the relative Dirichlet Energy of these uncontextualized tokens to the random walk tokens.

\section{Metalearning AWM Prompt}
\label{App:metalearning}
\begin{figure}
    \centering
    \includegraphics[width=\linewidth]{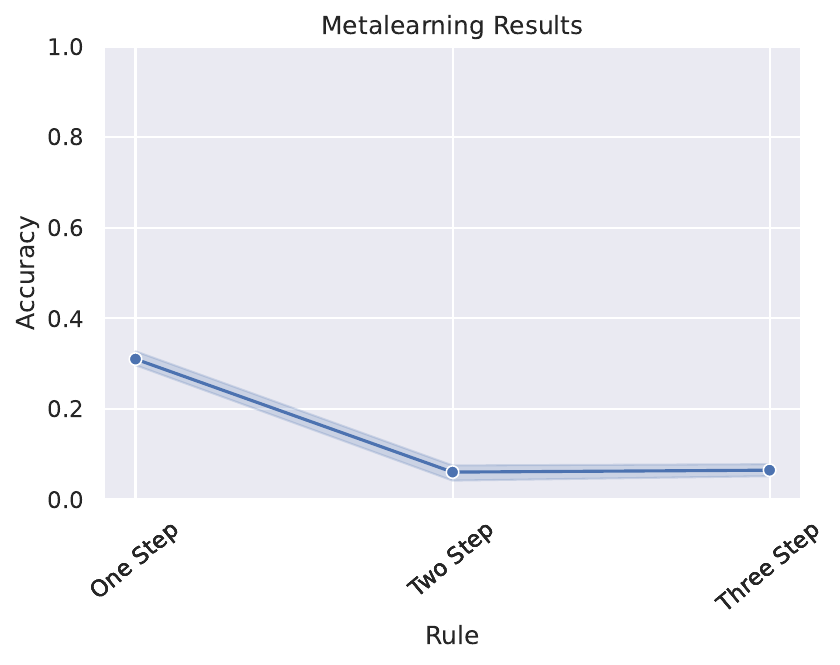}
    \caption{Metalearning prompt accuracy.}
    \label{fig:metalearning}
\end{figure}

In this section, we present the results of a metalearning version of the task, mentioned in Section~\ref{Sec:Exp2}. This variant consists of three task presentations, where each task presentation consists of a 500-length random walk over a 5-by-5 grid state space, followed by 3 few shot examples (all generated using the same rule). After these three task presentations, a query input is presented, and the model must infer the correct output. All task presentations are created using identical topologies, with the same tokens assigned to each state. 
These results are generated using a 5-by-5 grid topology. From Figure~\ref{fig:metalearning}, we see that this approach yields poor performance.

\section{Long Context Adaptive World Modeling}
\label{App:Long_ICWM}
\begin{figure}
    \centering
    \includegraphics[width=\linewidth]{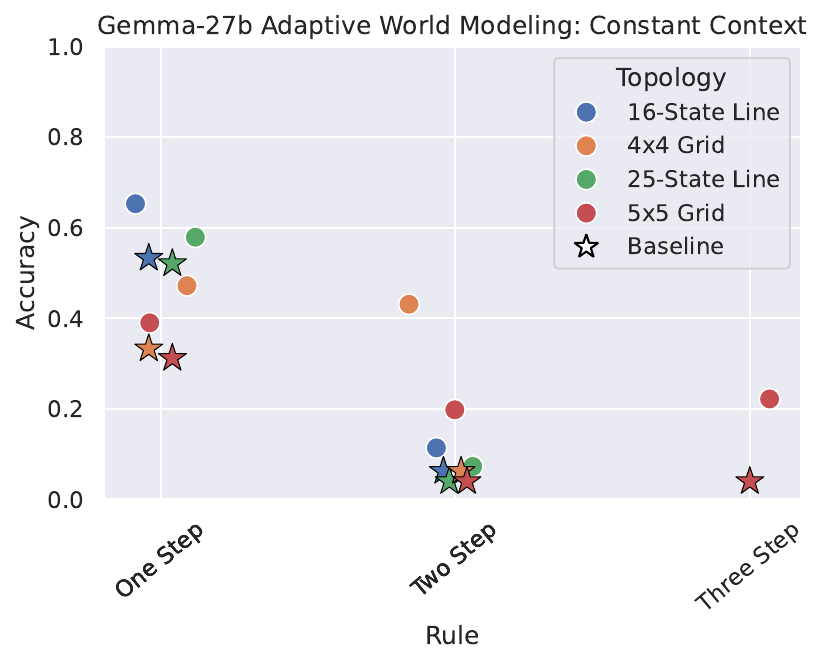}
    \caption{adaptive world modeling results with constant long context length, using \texttt{gemma-3-27b-it}.}
    \label{fig:icwm_cc}
\end{figure}

In this section, we present a similar analysis to Appendix~\ref{App:Long_NTP}, except for the adaptive world modeling task using \texttt{Gemma-3-27b-it}. From Figure~\ref{fig:icwm_cc}, we see that this does not improve performance.

\section{Frontier Model Next-Token Prediction Results}
\label{App:Frontier_NTP}
\begin{figure}
    \centering
    \includegraphics[width=\linewidth]{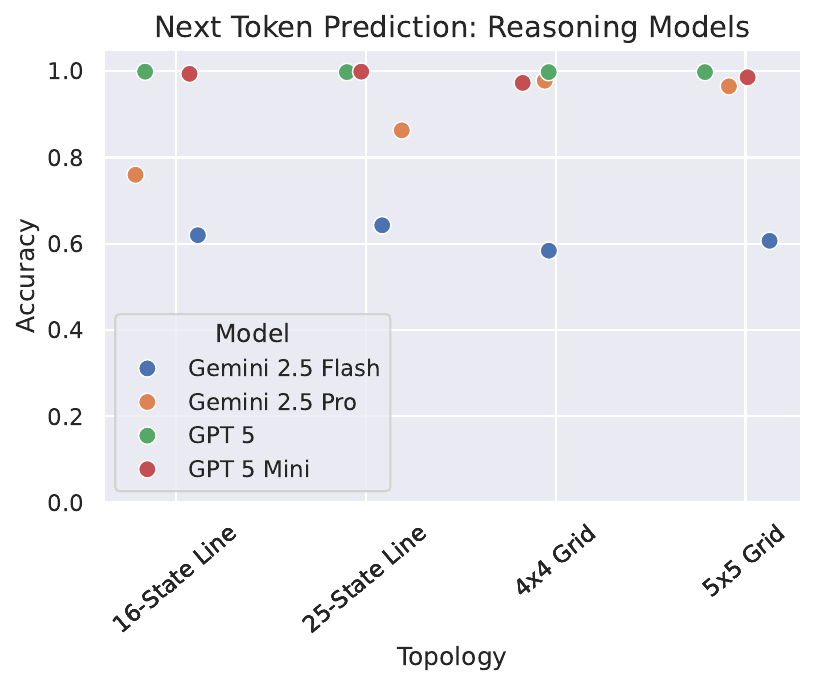}
    \caption{Frontier model next-token prediction performance.}
    \label{fig:app_frontier_ntp}
\end{figure}

In this section, we present results on the next-token prediction task using the frontier models described in Section~\ref{Sec:Exp3}. From Figure~\ref{fig:app_frontier_ntp}, we find that, aside from \texttt{Gemini-2.5-Flash}, models achieve good performance at next-token prediction. Surprisingly, this is especially true for grid topologies.

\section{Ablating Thinking}
\label{App:Ablating_Thinking}

\begin{figure}
    \centering
    \includegraphics[width=\linewidth]{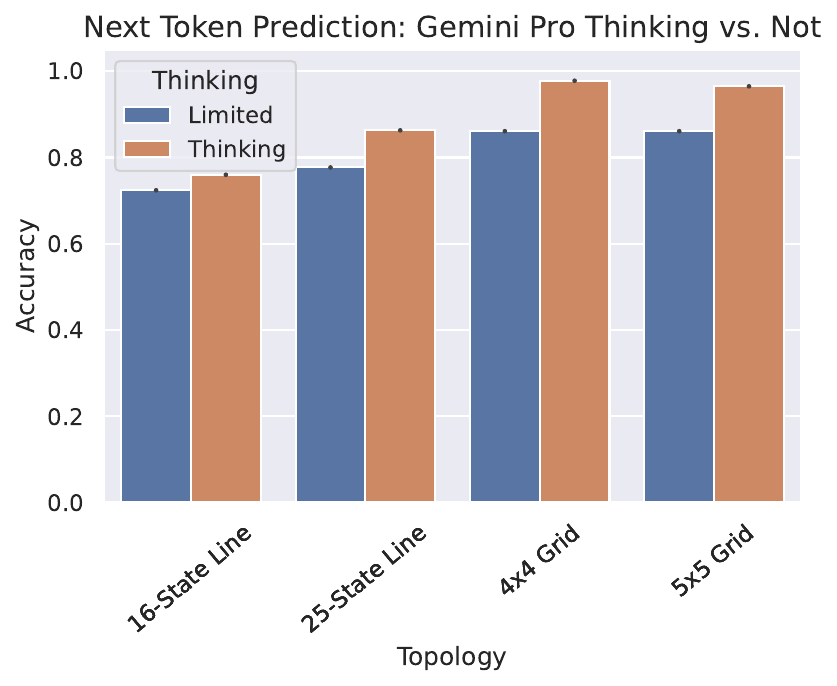}
    \caption{\texttt{Gemini-2.5-Pro} Thinking ablation for next-token prediction.}
    \label{fig:app_pro_ntp_ablation}
\end{figure}

\begin{figure}
    \centering
    \includegraphics[width=\linewidth]{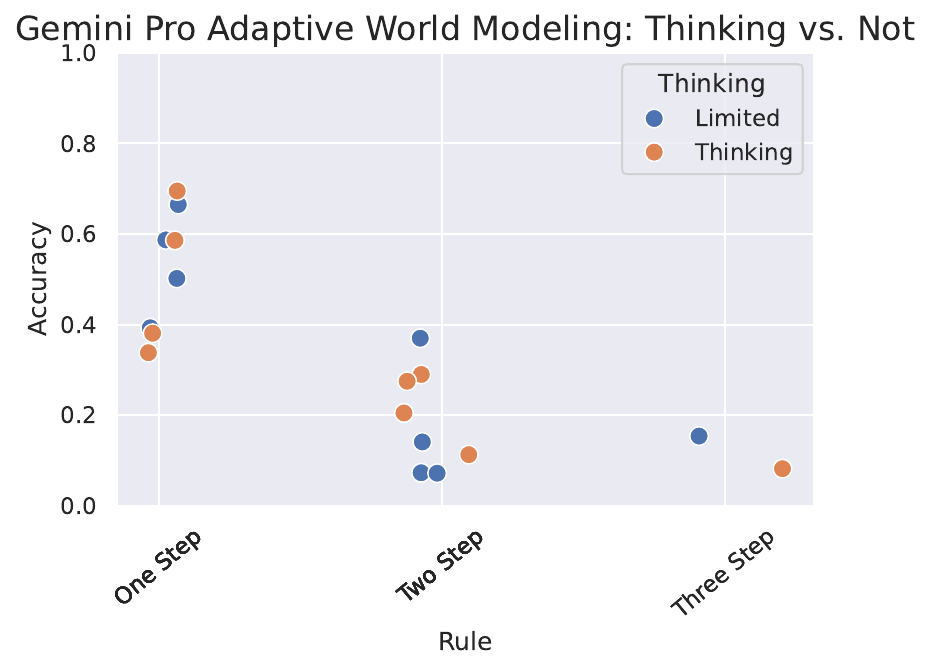}
    \caption{\texttt{Gemini-2.5-Pro} Thinking ablation for adaptive world modeling.}
    \label{fig:app_pro_icwm_ablation}
\end{figure}

\begin{figure}
    \centering
    \includegraphics[width=\linewidth]{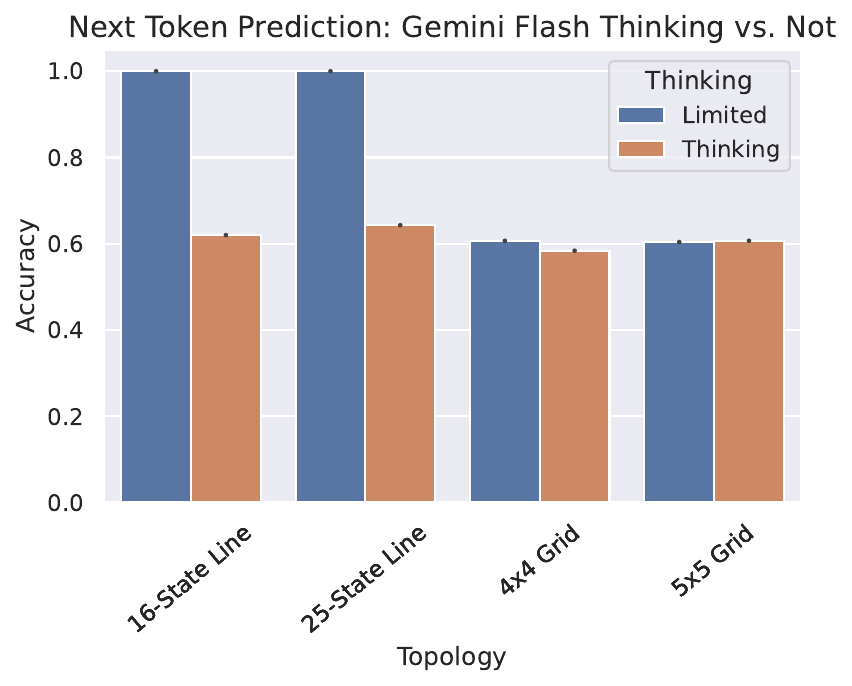}
    \caption{\texttt{Gemini-2.5-Flash} Thinking ablation for next-token prediction.}
    \label{fig:app_flash_ntp_ablation}
\end{figure}

\begin{figure}
    \centering
    \includegraphics[width=\linewidth]{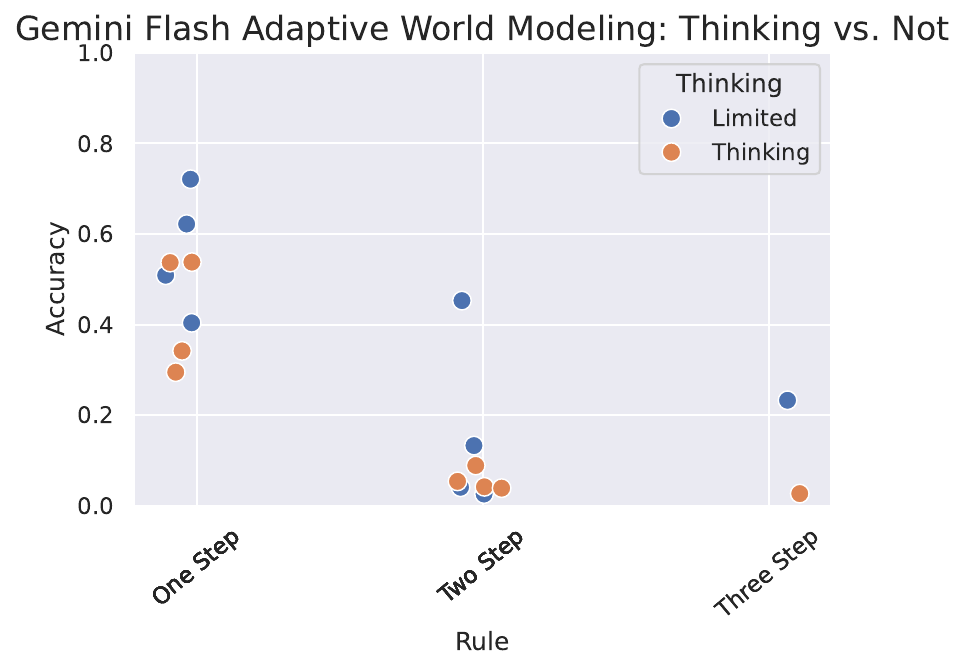}
    \caption{\texttt{Gemini-2.5-Flash} Thinking ablation for adaptive world modeling.}
    \label{fig:app_flash_icwm_ablation}
\end{figure}

In this section, we ablate the \texttt{Gemini} model's ability to use thinking tokens and analyze the impact on next-token prediction and adaptive world modeling. For \texttt{Gemini-2.5-Flash}, we can entirely ablate the model's ability to produce thinking tokens. However, for \texttt{Gemini-2.5-Pro}, we can only set the thinking budget to a minimum of 128 tokens. For \texttt{Pro}, we see that limiting thinking to 128 tokens often hurts performance on both tasks (See Figures~\ref{fig:app_pro_ntp_ablation} and \ref{fig:app_pro_icwm_ablation}). For \texttt{Flash}, however, we see mixed results. In many cases (and especially for next-token prediction on line topologies), ablating thinking drastically \textit{improves} performance (See Figures~\ref{fig:app_flash_ntp_ablation} and \ref{fig:app_flash_icwm_ablation}). More work must be done to understand the implications of these results. One speculative hypothesis is that entirely ablating thinking forces the model to rely on forward pass latent representations, rather than (potentially faulty) externalized thinking traces. These latent representations have very likely undergone in-context representation learning, and (perhaps) the model has learned to flexibly use these representations in this very specific, limited context.



\section{The Impact of Hints}
\label{App:Hints}
\begin{figure}
    \centering
    \includegraphics[width=\linewidth]{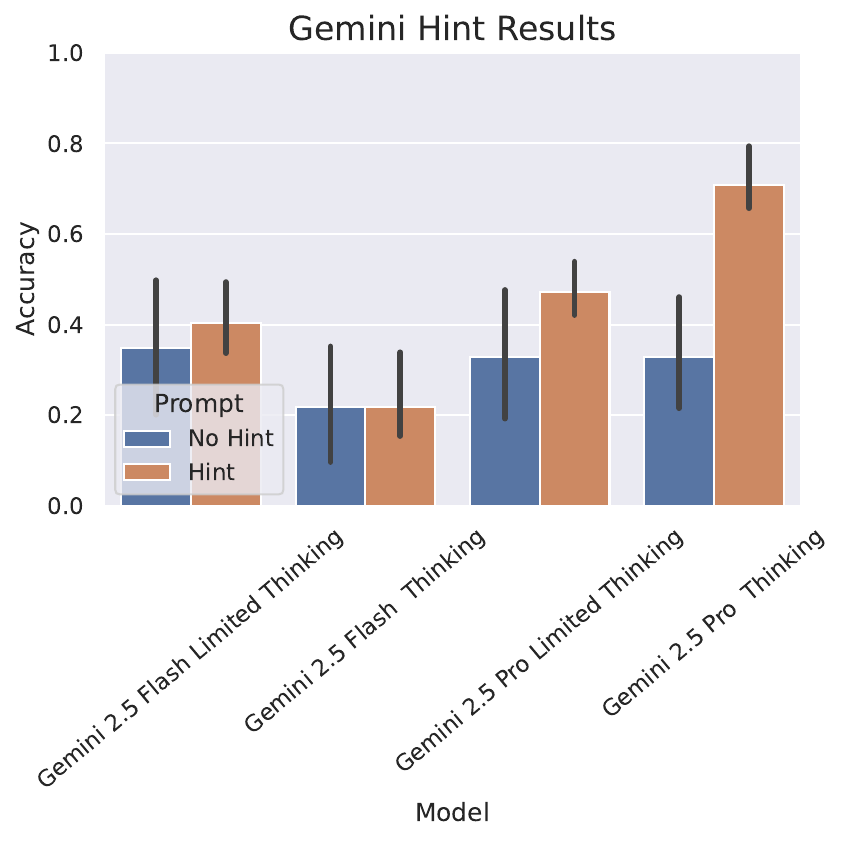}
    \caption{Adaptive world modeling performance on frontier models with and without hints on 2-dimensional topologies.}
    \label{fig:app_hints}
\end{figure}

In this section, we analyze frontier model performance when we provide strong clues about the underlying state space topology. In particular, here we include information about the configuration of the grid (4-by-4 or 5-by-5) in the prompt, and also mention that the few-shot examples map states at one position to states at another position. Even with these very strong hints, models still do not achieve perfect adaptive world modeling accuracy (See Figure~\ref{fig:app_hints}).

\end{document}